\def\year{2022}\relax
\documentclass[letterpaper]{article} 
\usepackage{aaai22}  
\usepackage{times}  
\usepackage{helvet}  
\usepackage{courier}  
\usepackage[hyphens]{url}  
\usepackage{graphicx} 
\usepackage{natbib}  
\usepackage{caption} 
\DeclareCaptionStyle{ruled}{labelfont=normalfont,labelsep=colon,strut=off} 
\usepackage{algorithm}
\usepackage{algorithmic}

\usepackage{makecell}
\usepackage{color}
\usepackage{graphicx}

\usepackage{stfloats}

\usepackage{newfloat}
\usepackage{listings}
\floatstyle{ruled}
\newfloat{listing}{tb}{lst}{}
\floatname{listing}{Listing}

\title{Impact of Color Variation on Robustness of Deep Neural Networks}

\author {
    Chengyin Hu \textsuperscript{\rm 1},
    Weiwen Shi \textsuperscript{\rm 1}
}
\affiliations {
    \textsuperscript{\rm 1} University of Electronic Science and Technology of China\\
    cyhuuestc@gmail.com, Weiwen\_shi@foxmail.com
}

\usepackage{bibentry}

\begin{document}

\maketitle

\begin{abstract}

Deep neural networks (DNNs) have have shown state-of-the-art performance for computer vision applications like image classification, segmentation and object detection. Whereas recent advances have shown their vulnerability to manual digital perturbations in the input data, namely adversarial attacks. The accuracy of the networks is significantly affected by the data distribution of their training dataset. Distortions or perturbations on color space of input images generates out-of-distribution data, which make networks more likely to misclassify them. In this work, we propose a color-variation dataset by distorting their RGB color on a subset of the ImageNet with 27 different combinations. The aim of our work is to study the impact of color variation on the performance of DNNs. We perform experiments on several state-of-the-art DNN architectures on the proposed dataset, and the result shows a significant correlation between color variation and loss of accuracy. Furthermore, based on the ResNet50 architecture, we demonstrate some experiments of the performance of recently proposed robust training techniques and strategies, such as Augmix, revisit, and free normalizer, on our proposed dataset. Experimental results indicate that these robust training techniques can improve the robustness of deep networks to color variation.

\end{abstract}

\section{Introduction}

Since the advent of AlexNet \cite{ref1} in 2012,  deep neural networks have demonstrated excellent performance on computer vision applications, such as image classification and object detection. Due to their high accuracy and excellent scalability on large-scale datasets, they are widely used in day-to-day life. However, recent studies have shown that they are vulnerable to human-designed variations or perturbations in the input data, which raises concerns about the robustness and security of deep neural networks. To mitigate the impact of the perturbations on the deep networks, researchers try to find more robust network architectures, and improve them with adversarial training \cite{ref2} and other strategies \cite{ref43,ref44}.

The architecture of deep neural networks is one of the key factors influencing their robustness. The development of network architectures began with AlexNet \cite{ref1}, which was one of the first deep neural networks to learn the semantic information with convolutional layer. It won the first place in the 2012 Imagenet \cite{ref3} image classification challenge, outperforming many traditional machine algorithms. The success of AlexNet started a boom in deep neural networks, motivating subsequent researchers to devote to the architectures of neural network. VGG \cite{ref4} replaced the convolutional kernels of AlexNet with smaller ones, which not only reduces the computational cost but also improves the accuracy performance of the network. Szegedy et al. \cite{ref5,ref23} proposed a module called inception in GoogleNet to simulate sparse networks with dense construction. Light weight architectures like Mobilenets \cite{ref7} provided trade-off between model size and accuracy. ResNet \cite{ref8} used residual modules to reduce the difficulty of learning identity maps,  solving the degradation problem of deep networks.

Recently, the vulnerability of deep networks has attracted significant attention \cite{ref9,ref10}. It was found that the networks are vulnerable to adversarial examples, namely input data with tiny and imperceptible perturbations, such as random noises and universal perturbations \cite{ref11,ref12}. However, most of related work have mainly focused on the impact of tiny, pixel-level and imperceptible perturbations on the classification results, while few studies the impact of global, geometric and structure transformation \cite{ref33}. In this work, we have studied the impact of color variation of the input images on the models’ performance.  It is well known that the performance of deep networks is highly relevant to the data distribution of its training dataset. Whereas if there is a shift in the distribution of the testing data from the training data, the accuracy of networks would drop, even if the semantic information of the input image stay unchanged. Shifts on the color channels of input image can change its data distribution, making the networks tend to output incorrect results. However, very few studies have studied the impact of color 
variation in images, even in the common image classification tasks. 

Color images contain more visual information than a gray-scale image and is more frequently used for extracting information in high-level computer vision tasks. Although color plays an important role in the visual information of the images, to the best of our knowledge no work has been able to explain how deep networks perceive them. While the working mechanic is still unclear, recent work by Kantipudi et al. \cite{ref13} has propsed a color channel perturbation attack on VGG, Resnet, and Densenet architectures and  reduces their accuracy  by average  41\%.

The current robustness benchmarking datasets like Imagenet-C, providing out-of distribution with noise, blur, weather, cartoons, sketches distortions. Hendrycks et al. \cite{ref20} and Lau F \cite{re21} respectively proposed a challenging dataset, Imagenet-A and NAO, which consist of real-world unmodified natural adversarial examples that most famous deep neural networks fail. As far as we know, there is few dataset designed for study the influence of color on deep networks. To help understand the impact of color variation, we propose an image dataset with different color variation on the RGB channels of images, generated from a subset of the Imagenet challenge dataset.

The main contributions of this paper include the creation of a dataset related to color-variation images to understand their impact, and then analyse the performance of advanced deep network architectures on image classification task on the proposed dataset. Besides, we examine the relation between the robustness of networks on color variation and their depth. Finally, we used ResNet50 as an example to study the influence of robust learning techniques like Augmix \cite{ref38}, Revisiting \cite{ref39} and Normalizer Free \cite{ref40} on networks’ capability to resist the impact of color. The rest of the paper is organized as follows: Section (Background) presents background information related to the existing literature, and Section (Experiments and Methodology) presents how we constructed the dataset and provides details of the experiments, followed by the results and findings, Section (Discussion) explores some of the interesting experiments and model attention analyses brought about by our approach and finally the conclusion and some future work are given in Section (Conclusion).

\section{Background}

Here, we present a review of relevant literature as well as some background knowledge. The quality of training data of deep neural networks have a great impact on their performance. The currently accepted hypothesis is that neural networks learn feature representation as well as semantic information in the data distribution from their training data. However, when the images are affected by perturbations like geometric transformation, deletion, and blur, their data distribution will change, which increases the probability of neural network to misclassify images. Szegedy \cite{ref6} first came up with the concept of adversarial attack, suggesting that it is the the error amplification effect rather than nonlinearity  or over-fitting that makes attacks work. Dodge and Karam \cite{ref14} provide a brief outlook on how adversarial samples affect the performance of the DNNs, and studied how distortion and perturbations affect the classification paradigm of them. They performed experiments on the Imagenet dataset with Caffe Reference \cite{ref22}, VGG16 \cite{ref4} and GoogLeNet \cite{ref5} models. They proposed a mixture of experts-based ensemble methods,  which uses a weighted sum of expert models to improve the robustness of network to multiple  types of distortions. Dodge and Karam \cite{ref24,ref25} further  made a comparison between the ability of human and deep neural networks to classify distorted images. The result reveals that humans outperform neural networks with distorted images, even when these networks are retrained with distorted data. 

Zhou et al. \cite{ref28} showed the performance of deep networks to classify distorted images can be improved by fine-tuning and re-training. Borkar and Karam \cite{ref26} propose a criteria to to evaluate the effect of perturbations like Gaussian blur and additive noise on the activations of pre-trained convolutional filters, and rank the most noise vulnerable convolutional filters in the common used CNN  in order to gain the highest improvement in classification accuracy upon correction. Hossain et al. \cite{ref27} also analyzed the performance of VGG16 when influenced by different types of perturbations, such as Gaussian white noise, Scaling Gaussian noise, salt \& pepper noise, speckle, motion blur,  and Gaussian blur. They used  discrete cosine transform while training to improve the robustness facing above distortions.

\textbf{Impact of Color.} Although many work have studied the impact of quality of images on the performance of deep neural networks, little of them have focus on the color-related distortions. Dosovitskiy and Brox \cite{ref15} were the first to show that manipulating the color of an object in a way that deviates from the training data has a negative impact on classification performance. Engilberge et al. \cite{ref16} managed to identify the color-sensitive units that processed hue characteristics in the VGG-19 and AlexNet. Kantipudi et al. \cite{ref13} proposed a color channel perturbation attack to fool deep networks and defense it by data augmentation. Le and Kayal \cite{ref17} compared various models to show that the robustness of edge detection is an important factor contributing to the robustness of models against color noise. Kanjar et al. \cite{ref18} analyzed the impact of color on robustness of widely used deep networks. They performed experiments on their proposed dataset with hue space based distortion. We use the validation set of the Imagenet database as our base database and augment different color-variation images from these images. Their work inspire us to further study the impact of color variation on deep networks.

\begin{figure*}
\centering
\includegraphics[width=1\linewidth]{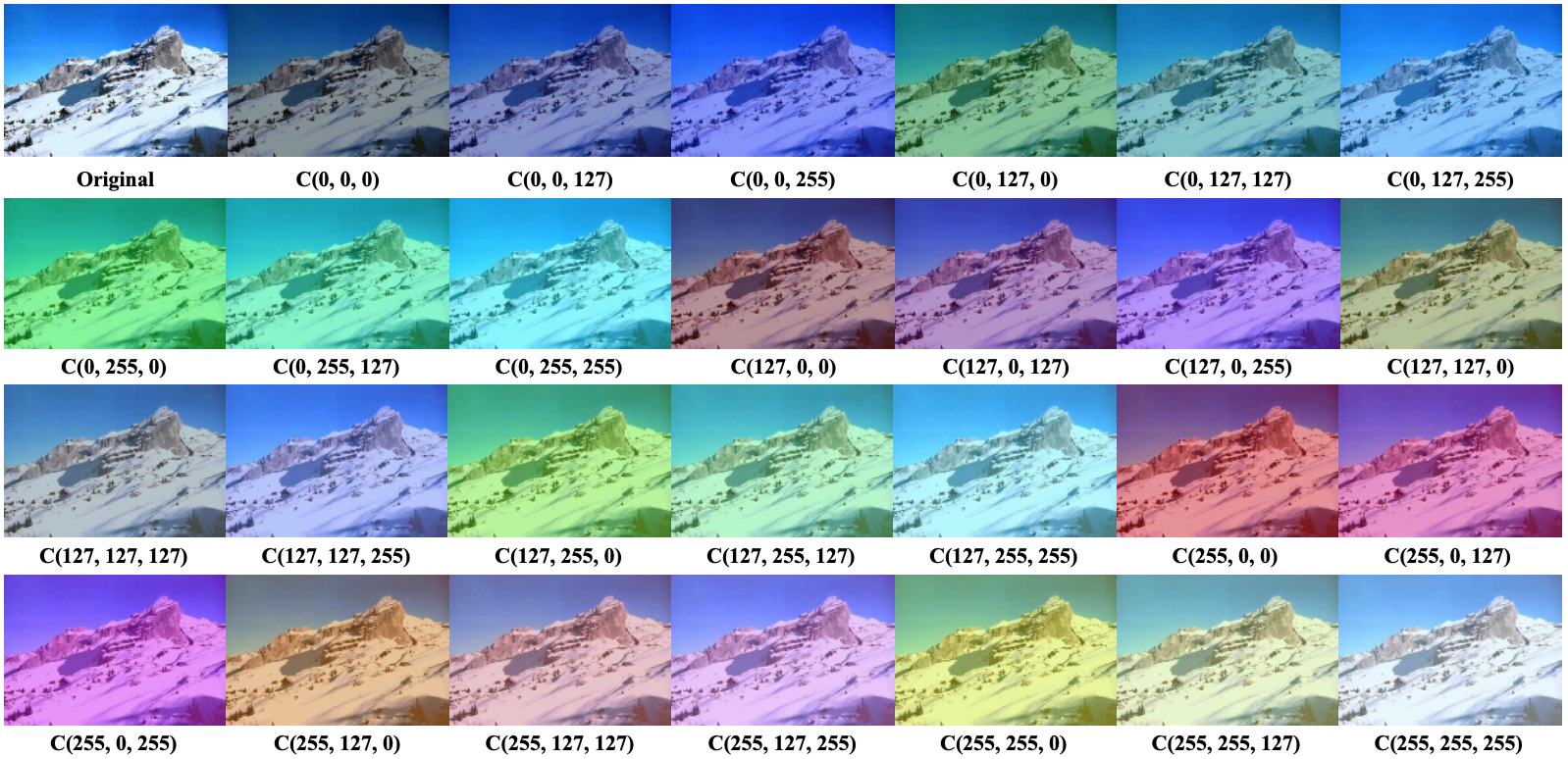} 
\caption{Examples of generating color-distortion images.}
\label{figure1}
\end{figure*}

\textbf{Architectures.} Alexnet \cite{ref1} is one of the origins of nowadays most common neural network architectures. It pioneered  to take the use of GPUs to accelerate the training of neural networks, reducing the training time of the neural network to an acceptable range. The success of AlexNet on Imagenet has motivated more work on the development of DNNs’ architecture. VGG \cite{ref4} took advantage of Alexnet and  used several consecutive $3\times3$ convolution kernels instead of the larger ones in AlexNet ($11\times11$, $7\times7$, $5\times5$) to improve the performance. They pointed out that for a given receptive field, using consecutive smaller convolution kernels is better than a larger convolution kernel, as it makes the network deeper and more efficient. Another key innovations in deep network architectures is the inception module, which consists of  $1\times1$, $3\times3$ and $5\times5$ convolutions. Inception approximate and covers the best local sparse structure of the convolutional network by easily accessible dense components. With the rapid development of the architecture of deep networks,  the models have become much deeper, which caused vanishing gradients and degradation problems. The ResNet \cite{ref8} architecture was proposed to tackle these issues, and is still one of the widely used backbones in computer vision tasks nowadays. ResNet proposed a structure called residual block, which skips connections between adjacent layers, enabling the network to learn identity mapping easier. It ensures that the deeper network at least perform as good as smaller ones.  Based on ResNet architecture, DenseNet \cite{ref41} proposes a more radical intensive connectivity architecture, which connects all the layers to each other and each layer takes all the layers before it as its input. DenseNet needs fewer parameters compared to the other traditional convolutional neural networks by reducing the need to learn redundant features. As the networks become deeper, their drawbacks of being computationally intensive and require a lot of GPU memory become more significant, which makes them unsuitable for mobile devices. To deploy models in such portable devices a group of lightweight networks were proposed, and Mobilenet \cite{ref7} is one of the most famous architectures among them. It proposed the concepts of depth wise separable convolutions and inverted residuals, which achieve similar performance to traditional networks with less computational cost.

\begin{table*}[h] 
    \centering
    \caption{\label{Table 1} Classification Top-1 accuracy (\%) of well-known networks on the Imagenet-CV dataset.}
    \begin{tabular}{ccccccc}
    \quad & DenseNet & ResNet50 & VGG19 & GoogleNet & MobileNet & AlexNet\\
    \hline
    
    Original & 82.14 & 85.83 & 82.51 & 75.87 & 80.70 & 71.59\\
    \hline
    
    C(0, 0, 0) & 80.83 & 82.65 & 79.25 & 74.04 & 76.64 & 64.04\\
    \hline
    C(0, 0, 127) & 75.59 & 75.59 & 70.38 & 70.70 & 67.86 & 44.22\\
    \hline
    C(0, 0, 255) & 67.86 & 64.97 & 57.35 & 65.42 & 54.99 & 22.47\\
    \hline
    C(0, 127, 0) & 70.57 & 68.82 & 63.46 & 70.41 & 59.87 & 29.31\\
    \hline
    C(0, 127, 127) & 73.54 & 73.67 & 65.76 & 69.62 & 64.03 & 36.17\\
    \hline
    C(0, 127, 255) & 70.48 & 69.56 & 59.39 & 64.67 & 57.73 & 30.20\\
    \hline
    C(0, 255, 0) & 62.68 & 51.72 & 52.53 & 66.44 & 43.81 & 11.60\\
    \hline
    C(0, 255, 127) & 65.20 & 61.11 & 56.71 & 66.84 & 49.22 & 17.04\\
    \hline
    C(0, 255, 255) & 64.51 & 63.41 & 54.82 & 63.95 & 51.00 & 18.53\\
    \hline
    
    C(127, 0, 0) & 73.91 & 72.97 & 68.42 & 71.34 & 65.37 & 40.06\\
    \hline
    C(127, 0, 127) & 72.58 & 70.52 & 66.68 & 71.68 & 63.13 & 35.23\\
    \hline
    C(127, 0, 255) & 65.63 & 58.16 & 55.58 & 67.83 & 55.63 & 19.70\\
    \hline
    C(127, 127, 0) & 74.74 & 75.15 & 69.56 & 71.75 & 67.24 & 42.06\\
    \hline
    C(127, 127, 127) & 80.78 & 81.94 & 77.74 & 74.86 & 75.17 & 60.61\\
    \hline
    C(127, 127, 255) & 75.26 & 75.63 & 68.12 & 71.05 & 67.19 & 41.55\\
    \hline
    C(127, 255, 0) & 63.26 & 58.72 & 53.44 & 67.61 & 48.35 & 15.85\\
    \hline
    C(127, 255, 127) & 69.55 & 69.11 & 60.52 & 70.63 & 57.78 & 26.15\\
    \hline
    C(127, 255, 255) & 72.03 & 72.61 & 62.08 & 69.50 & 60.81 & 30.84\\
    \hline
    
    C(255, 0, 0) & 66.78 & 62.80 & 57.81 & 67.02 & 53.25 & 24.63\\
    \hline
    C(255, 0, 127) & 67.74 & 60.56 & 59.12 & 66.37 & 51.73 & 20.84\\
    \hline
    C(255, 0, 255) & 62.93 & 49.27 & 48.98 & 67.30 & 44.44 & 12.07\\
    \hline
    C(255, 127, 0) & 71.38 & 71.20 & 65.77 & 68.97 & 62.74 & 36.68\\
    \hline
    C(255, 127, 127) & 73.42 & 73.91 & 68.14 & 71.66 & 65.71 & 38.32\\
    \hline
    C(255, 127, 255) & 70.98 & 68.07 & 61.81 & 71.52 & 60.85 & 28.95\\
    \hline
    C(255, 255, 0) & 65.76 & 63.81 & 55.13 & 68.12 & 54.22 & 23.35\\
    \hline
    C(255, 255, 127) & 73.70 & 73.82 & 66.41 & 71.64 & 64.44 & 38.71\\
    \hline
    C(255, 255, 255) & 79.50 & 79.44 & 73.52 & 73.94 & 71.48 & 52.35\\
    \hline
    
    \end{tabular}
\end{table*}

\textbf{Robustness.} Evaluating robustness of deep neural networks is still an challenging and ongoing area of research \cite{ref29}. Some works use data augmentation to improve the robustness of networks \cite{ref34}. Papernot \cite{ref35} et al. first pointed out some limitations of deep learning in adversarial settings. They proposed forward derivative attack to fool deep networks by only alter a minority of input. Hendrycks et al. \cite{ref36} defined some benchmark metrics of robustness of deep neural network to some common perturbations like additive noise, blur, compression artifacts, etc. They proposed a variant of Imagenet referred to as Imagenet Challenge(or Imagenet-C). Imagenet-C contains 15 types of automated generated perturbations, on which many well-known deep networks perform poorly. Kanjar et al. \cite{ref18} analyzed the impact of color on robustness of widely used deep networks. Recent studies have indicated that deep convolutional neural networks pre-trained on Imagenet dataset are vulnerable to texture bias \cite{ref38}, while the impact of color variation in images is not deeply studied. Xiao et al. \cite{ref31} have formalized the scaling attack, illustrating its goal, generation algorithms, and optimization solution. Zheng et al. \cite{ref32} provide a basis for robustness evaluation and conduct experiments in different situations to explore the relationship between image scaling and the robustness of adversarial examples. With the development of  adversarial attack techniques, many studies focus on defensing against these attacks and try to find feasible training strategies to improve the robustness of models \cite{ref19}. Augmix \cite{ref38} is a simple training strategies which uses several augmentation techniques together with Jenson-Shannon divergence loss to enforce a common embedding for the classifier. Brock et al. \cite{ref40} proposed a normalized family of free networks called NF-Nets to prevent the gradient explosion by not using batch normalization. \cite{ref39} recently showed that network models can effectively improve the classification performance of ResNet models by using some scaling strategies and developed a set of models called ResNet-RS.

\section{Experiments and Methodology}

Here, we present the details of data generation process of our proposed dataset and the neural network architecture used in our experiments.

\subsection{Dataset Generation}

Typically, DNNs trained on Imagenet dataset have 1000 different labels. Our proposed dataset is derived from a subset of Imagenet Challenge dataset, and we call it Imagenet-ColorVariation (Imagenet-CV). Firstly, 50 images were randomly selected from each class of Imagenet Challenge to generate a clean sample dataset, it totally contains 50,000 raw images. Secondly, the whole dataset of 1,350,000 images is generated by adding 27 colors to each raw image respectively with transparency of 0.4. As shown in Figure \ref{figure1}, one original images was augmented into 27 images with different color variation.

\subsection{Impact on widely used network architectures}
In this section, we perform some experiments on the accuracy of six widely used deep network architectures \cite{ref41,ref8,ref4,ref23,ref7,ref1} on the proposed Imagenet-CV dataset. In the table, the first row(Original) represents the classification accuracy of the images in the absence of any color perturbation, and the others were distorted with corresponding RBG values on their color channels.

As can be seen from the experimental results in Table \ref{Table 1}, different color variation decreases the classification top-1 accuracy of advanced DNNs. This observation is consistent throughout all architectures. When further investigate the influence of different architectures, it is found that the ResNet architectures perform best among these models. DenseNet and VGG19 architectures show similar classification accuracy and perform better than MobileNet and GoogleNet. And AlexNet show the worst classification performance in each color variation. 

\begin{figure*}
\centering
\includegraphics[width=1\linewidth]{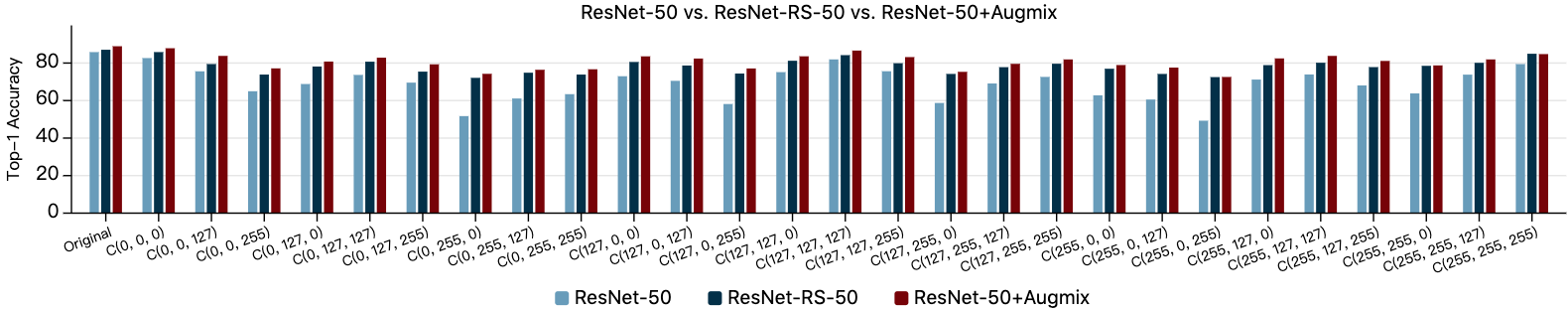} 
\caption{Performance of ResNet50 vs. ResNet-RS-50 vs. ResNet50+Augmix.}
\label{figure2}
\end{figure*}

\begin{figure*}[b]
\centering
\includegraphics[width=1\linewidth]{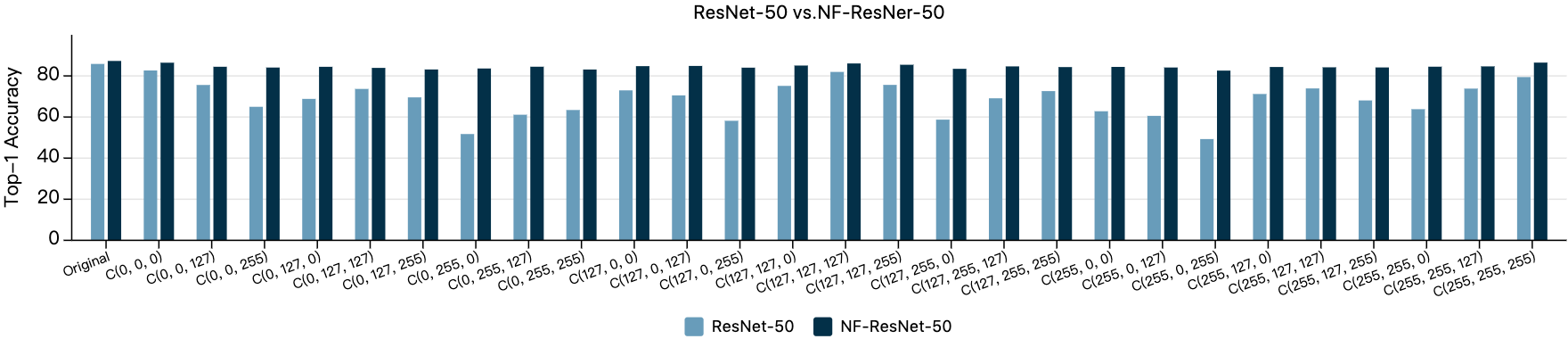} 
\caption{Performance of ResNet50 vs. NF-ResNet50.}
\label{figure3}
\end{figure*}

It indicates that although the visual information of the images are barely changed, the color variation on images still have significant impact on the performance deep neural networks. Despise the tested deep networks all contains convolutional layers, which gives them the property of geometric invariance, they are still vulnerable to color variation. Another interesting insights in the result is that when we look deeper into the results, we can find that adding green channel values $C(0, 255, 0)$ can make more accuracy decline compared to blue and red channels. We infer that this phenomenon suggest there are too many green color pictures in the training set. So when we add green value to the input data, they are more likely to allow the networks to misclassification.

\subsection{Recent advances in efficient and robust models}
In this section we introduce some advances of robust training strategies, and show the performance of Resnet models trained with these strategies on our proposed dataset.

\begin{table*}[h]
    \centering
    \caption{\label{Table 2} Performance of ResNet models with different depth.}
    \begin{tabular}{cccccc}
    \quad & ResNet18 & ResNet34 & ResNet50 & ResNet101 & ResNet152\\
    \hline
    
    Original & 77.90 & 83.21 & 85.83 & 88.51 & 88.92\\
    \hline
    
    C(0, 0, 0) & 62.88 & 69.57 & 71.96 & 75.54 & 77.68\\
    \hline
    C(0, 0, 127) & 52.75 & 57.31 & 60.29 & 65.65 & 68.87\\
    \hline
    C(0, 0, 255) & 41.00 & 43.51 & 48.87 & 57.72 & 62.87\\
    \hline
    C(0, 127, 0) & 54.36 & 60.33 & 64.68 & 67.78 & 70.84\\
    \hline
    C(0, 127, 127) & 56.27 & 60.57 & 65.18 & 68.59 & 71.70\\
    \hline
    C(0, 127, 255) & 50.59 & 51.38 & 57.65 & 62.72 & 67.01\\
    \hline
    C(0, 255, 0) & 42.99 & 47.12 & 52.16 & 58.37 & 61.92\\
    \hline
    C(0, 255, 127) & 42.97 & 45.89 & 52.47 & 57.86 & 62.71\\
    \hline
    C(0, 255, 255) & 41.38 & 42.79 & 49.55 & 56.34 & 61.67\\
    \hline
    
    C(127, 0, 0) & 53.57 & 57.54 & 61.12 & 67.41 & 69.62\\
    \hline
    C(127, 0, 127) & 53.07 & 56.48 & 58.20 & 64.80 & 66.38\\
    \hline
    C(127, 0, 255) & 43.09 & 46.95 & 49.42 & 59.28 & 63.55\\
    \hline
    C(127, 127, 0) & 58.54 & 64.65 & 67.13 & 71.95 & 73.34\\
    \hline
    C(127, 127, 127) & 68.57 & 73.77 & 77.31 & 79.95 & 81.74\\
    \hline
    C(127, 127, 255) & 56.22 & 59.59 & 63.78 & 69.00 & 71.12\\
    \hline
    C(127, 255, 0) & 44.22 & 50.18 & 52.35 & 58.19 & 61.02\\
    \hline
    C(127, 255, 127) & 49.44 & 53.40 & 59.10 & 62.61 & 65.85\\
    \hline
    C(127, 255, 255) & 50.15 & 51.58 & 58.04 & 62.07 & 65.83\\
    \hline
    
    C(255, 0, 0) & 43.75 & 46.21 & 51.83 & 60.86 & 63.38\\
    \hline
    C(255, 0, 127) & 44.54 & 47.04 & 50.71 & 59.40 & 61.92\\
    \hline
    C(255, 0, 255) & 37.51 & 41.55 & 45.33 & 56.60 & 58.91\\
    \hline
    C(255, 127, 0) & 47.68 & 50.76 & 55.34 & 62.34 & 65.45\\
    \hline
    C(255, 127, 127) & 54.67 & 57.36 & 60.52 & 67.20 & 69.36\\
    \hline
    C(255, 127, 255) & 46.15 & 49.28 & 51.69 & 59.80 & 61.21\\
    \hline
    C(255, 255, 0) & 36.22 & 42.84 & 42.97 & 52.78 & 56.47\\
    \hline
    C(255, 255, 127) & 49.38 & 53.96 & 56.14 & 61.84 & 65.50\\
    \hline
    C(255, 255, 255) & 56.68 & 60.25 & 64.14 & 67.53 & 71.38\\
    \hline
    
    \end{tabular}
\end{table*}

\textbf{Augmix and ResNet-RS-50.} Hendrycks et al. \cite{ref38}  propsed Augmix as s a data processing technology  to improve the robust performance of DNNs. The augmentation of this technique includes rotation, translation, separation and other enhancement technologies. It achieves simple data processing within limited computational overhead, helping the model to withstand unforeseen corruptions, and Augmix significantly improves robustness and uncertainty in challenging image classification benchmarks. Here we show the classification performance of ResNet50 with Augmix data processing technology on our proposed Imagenet-CV dataset. We compared the classification performance of pretrained ResNet50 model with that of pretrained ResNet50 model with Augmix technology, and the classification comparison is shown in Figure \ref{figure2}. It can be seen that when the input images is distorted in color channels, the classification performance of ResNet50 with Augmix technology is better than that of ordinary pretrained ResNet50 model. Bello et al. \cite{ref39} recently showed that scaling network models can effectively improve the classification performance of models, and developed a set of models called ResNet-RS. It pointed out that the training and extension strategy may be more important than the architecture changes, and the ResNet architecture designed with the improved training and extension strategy is 1.7-2.7 times faster than the EfficientNets on TPUs, while achieving similar accuracy on ImageNet. In the large-scale semi-supervised learning setting, ResNet-RS achieves 86.2\% Top-1 ImageNet accuracy while being 4.7 times faster than EfficientNet-NoisyStudent. Here, we show the classification performance of ResNet-RS-50 on Imagenet-CV and its Top-1 accuracy is shown in Figure \ref{figure2}. It can be seen that ResNet-RS-50 has better robustness than pretrained ResNet50.

In general, the robustness of ResNet50 with Augmix and ResNet-RS-50 are better than the original pretrained ResNet50 model. In addition, it can be seen from Figure \ref{figure2} that, when the color is $C(0, 255, 255)$ and $C(255, 0, 255)$, ResNet50 has the lowest classification accuracy, while ResNet-RS-50 and ResNet50+Augmix still maintain a certain robustness.

\textbf{Normalizer Free ResNet50(NF-ResNet50).} Brock et al. \cite{ref40} proposed a family of normalized-free networks called NF-Nets, which abandoned the traditional concept that data need to be normalized and did not use batch normalization, achieving the best level in the industry on large image classification tasks. They proposed Adaptive Gradient Clipping methods to realize non-normalized networks augmented with larger quantities of subscale and large scale data. In the training of NF-Nets, the gradient size is limited to effectively prevent gradient explosion and training instability. experiments have shown the superiority of their method. Figure \ref{figure3} shows the Top-1 accuracy of NF-ResNet50 on Imagenet-CV. It can be seen that, compared with pretrained ResNet50 model, the classification performance of NF-ResNet50 is much higher at different color variation.

\subsection{ResNet models with different depths}

Before the advent of ResNet, the neural networks were not very deep, as they were suffer from the gradient disappearance or gradient explosion problems. VGG’s network has only 19 layers, while ResNet had a whopping 152 layers. Many people have the intuitive impression that the more layers of the network, the better the training effect. However, the accuracy of the training model is not necessarily correlated with the number of their layers. Because with the deepening of the network layer, the accuracy of network needs to appear saturation, and appear as the phenomenon of decline. Suppose a 56 network than 20 layer network training effect is poor, many people first reaction is overfitting, but this is not the case, because the accuracy of overfitting phenomenon of the training set will be very high, but 56 layer network training set accuracy may also is very low, it shows that the network depth increase may not guarantee the accuracy of classification. It is obvious that with the deepening of layers, there will be gradient disappearance or gradient explosion, which makes it difficult to train deep models. However, BatchNorm and other methods have been used to alleviate this problem. Therefore, how to solve the degra dation problem of deep networks is the next direction of neural networks development. ResNet is proposed to mitigate the problem of vanishing/exploding gradient. One of the major advantages of residual neural network is the ability to learn identity mapping. It uses skip connections to allow alternate shortcuts path for the gradient to flow through, which  ensures that the higher layer will perform at least as good as the lower layer.

We use ResNet models with different depths (ResNet18, ResNet34, ResNet50, ResNet101, ResNet152) to test the  classification top-1 accuracy on the proposed Imagenet-CV. It can be seen from Table \ref{Table 2} that the classification Top-1 accuracy of ResNet series models is proportional to the model depth. However, when facing additional color variation, the deeper networks won't show higher robustness, as their accuracy decline proportionally.

\begin{figure*}
\centering
\includegraphics[width=1\linewidth]{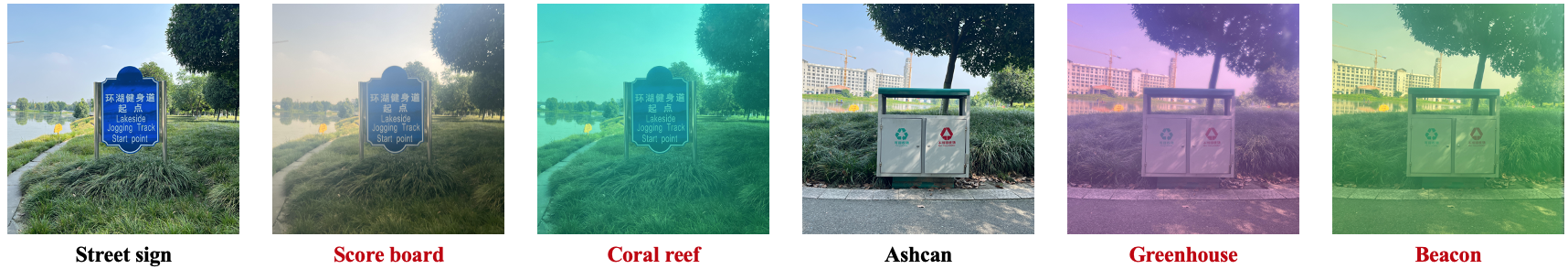} 
\caption{Adversarial samples in the physical world.}
\label{figure4}
\end{figure*}

\section{Discussion}
\subsection{Exploration of physical attacks}
In this section we conduct some interesting experiments to explore the feasibility of applying the proposed method to physical attacks.

We believe that color variation that mislead models in digital settings could also impact the performance of deep networks in physical scene. We photograph some clean samples in the physical world, and then add color distortions to the clean samples with the proposed method. In order to simulate the color variation to generate physical samples  in real world, we use customized translucent filters with specific color to cover the camera and take the photos.

We used ResNet50 as the baseline model for this experiment, with street signs and ashcan as experimental subjects. As can be seen from the experimental demonstration in Figure \ref{figure4}, after adding color distortion to the street sign, the classifier misclassifies it as score board, coral reef, etc. Therefore, we believe that the proposed method is feasible for physical attacks.

\subsection{CAM for images}

\begin{figure}
\centering
\includegraphics[width=1\linewidth]{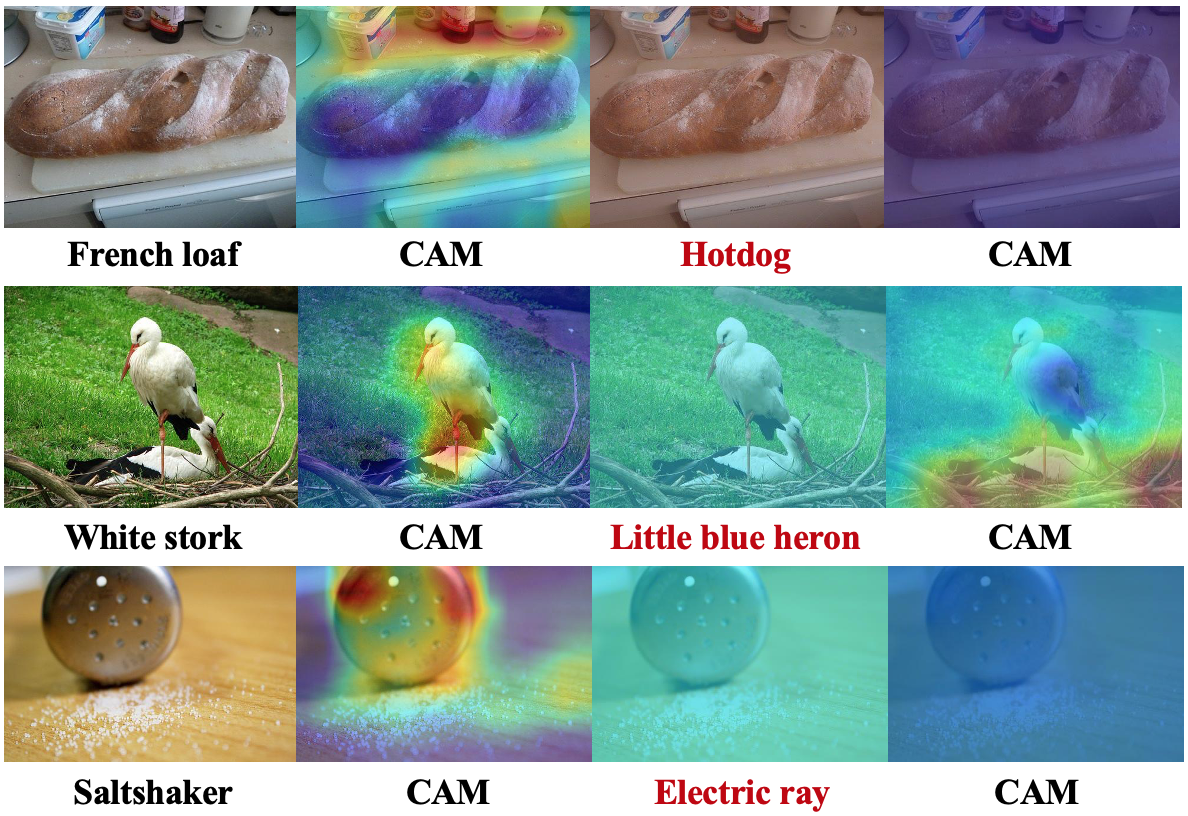} 
\caption{CAM for images.}
\label{figure5}
\end{figure}

From the sample demonstration in Figure \ref{figure1}, we can see that the semantic information of the sample is unchanged even when colors are added to the clean sample. We carefully check the misclassified images in the dataset, and then use CAM \cite{ref42} to show the model's attention. We found that when colors are added to images, the model's attention would disappear or shift from the target object to another one. As shown in Figure \ref{figure5}, the model's attention disappears when color is added to french loaf. Besides, when color is added to the white stork, the model's attention shifts to other objects.

\section{Conclusion and future work}
With deep neural networks being widely used in our daily life, it is crucial to study the robustness of deep learning models and try to make these models more robust and resistible to perturbations. Experimental studies presented in this paper have yielded some interesting results with respect to the impact of color variation of images on the performance of deep neural network architectures with respect to the shift in data distribution. The performance of these networks drastically reduces when adding various colors to the channels of images. In this paper, we have presented the overall classification accuracy performance of some widely used deep neural network architectures under different color variations and the results demonstrated will serve as a motivation to investigate the color perception mechanism of further architectures studies in the future. The analysis mentioned in this paper will motivate researchers to take into consideration the impact of color and aspects of other perturbations to propose more accurate and robust models based on deep neural networks. The important observations are listed as follows:

\begin{itemize}

\item 
There is a significant impact of color variation on the inference of deep neural networks. 
\item
Data processing and data augmentation techniques like Augmix have some positive impact on robustness and optimizing the training procedure of deep networks, making Resnet-RS-50 a much more robust model compared to Resnet-50 with respect to color variation. 
\item
Training procedures like adversarial prop and noisy student training offer some amount of additional robustness to models. 
\item
The Normalizer free models offer more robustness to color related transformation. 
\item
Deeper ResNet don’t significantly show more robustness compared to networks with fewer layers when facing color variation.

\end{itemize}


\bibliography{aaai22}

\end{document}